\pdfoutput=1
\documentclass[11pt]{article}

\usepackage[]{ACL2023}

\usepackage{times}
\usepackage{latexsym}

\usepackage[T1]{fontenc}
\usepackage[utf8]{inputenc}

\usepackage{microtype}

\usepackage{inconsolata}

\usepackage{amsthm}
\usepackage{amsmath}
\usepackage{amssymb}
\usepackage{booktabs}
\usepackage{graphicx}
\usepackage{caption}
\usepackage{subcaption}
\usepackage{multirow}
\usepackage{makecell}
\usepackage{url}
\usepackage{soul} 
\usepackage{xcolor}
\usepackage{pifont}

\usepackage{color}
\definecolor{Red}{RGB}{255,45,61}
\definecolor{Orange}{RGB}{237,125,49}
\definecolor{Green}{RGB}{0,176,80}
\definecolor{Blue}{RGB}{0,112,192}

\newtheorem{definition}{Definition}
\newcommand\nelname{NEL}
\newcommand\clinkname{CLINK}
\newcommand\nil{NIL}

\title{Learn to Not Link: Exploring \nil{} Prediction in Entity Linking}

\author{Fangwei Zhu$^{1,2}$, Jifan Yu$^{1,2}$, Hailong Jin$^{1,2}$, Juanzi Li$^{1,2}$, Lei Hou$^{1,2}$\thanks{\quad Corresponding Author}\hspace{0.5em} and Zhifang Sui$^{3}$ \\
  $^1$Dept. of Computer Sci.\&Tech., BNRist, Tsinghua University, Beijing 100084, China \\
  $^2$KIRC, Institute for Artificial Intelligence, Tsinghua University \\
  $^3$MOE Key Lab of Computational Linguistics, Peking University \\
  \texttt{\{zfw19@mails.,yujf18@mails,jinhl15@mails.,lijuanzi@,houlei@\}tsinghua.edu.cn} \\
  \texttt{\{szf@\}pku.edu.cn} \\
}

\begin{document}
\maketitle
\begin{abstract}
Entity linking models have achieved significant success via utilizing pretrained language models to capture semantic features.
However, the \nil{} prediction problem, which aims to identify mentions without a corresponding entity in the knowledge base, has received insufficient attention.
We categorize mentions linking to \nil{} into Missing Entity and Non-Entity Phrase, and propose an entity linking dataset \nelname{} that focuses on the \nil{} prediction problem.
\nelname{} takes ambiguous entities as seeds, collects relevant mention context in the Wikipedia corpus, and ensures the presence of mentions linking to \nil{} by human annotation and entity masking.
We conduct a series of experiments with the widely used bi-encoder and cross-encoder entity linking models, results show that both types of \nil{} mentions in training data have a significant influence on the accuracy of \nil{} prediction.
Our code and dataset can be accessed at \url{https://github.com/solitaryzero/NIL_EL}.
% The dataset and code will be released upon acception.
\end{abstract}

\section{Introduction}
Entity Linking (EL) aims to map entity mentions in free texts to their corresponding entities in a given Knowledge Base (KB).
Entity linking acts as a bridge between unstructured text and structured knowledge, and benefits various downstream tasks like question answering~\cite{luo2018knowledge} and knowledge extraction~\cite{chen2021lightweight}.

However, not all entity mentions correspond to a specific entity in the KB that suits the mention context~\cite{ling2015design}.
Take Table \ref{tab:nil_example} as an example, \textbf{Peter Blackburn} is actually a journalist, and \textbf{the householder} is a common phrase rather than an entity.
These two mentions do not refer to any entity in the given KB.
The identification of these mentions is referred to as \nil{} prediction.
Therefore, to tackle \nil{} prediction, the entity linking model needs to select mentions whose references are absent in KB, and link them to a special placeholder \textit{\nil{}}.
\citet{dredze2010entity} states \nil{} prediction as one of the key issues in entity linking, which may lead to a decrease in the recall of entity linking systems.
Meanwhile, the incorrectly linked entities may provide false information to downstream tasks.

There have been some earlier representation learning based researches that take \nil{} prediction into consideration~\cite{eshel2017named, lazic2015plato, peters2019knowledge}.
They identify \nil{} mentions by setting a vector similarity threshold or viewing \nil{} as a special entity.
Recently, pretrained language model (PLM) based models~\cite{wu2020scalable, fitzgerald2021moleman, decao2020autoregressive} have achieved great success for their great transferability and expandability.
However, these models generally assume that there always exists a correct entity for each mention in the knowledge base, which leaves the \nil{} prediction problem without adequate attention.

\sethlcolor{pink}

\begin{table*}[htbp]
    \centering
    \scalebox{0.9}{
    \begin{tabular}{p{0.18\linewidth}|p{0.7\linewidth}}
         \toprule
         \multicolumn{2}{c}{\textbf{Missing Entity}} \\
         \midrule
         Mention Context & EU rejects German call to boycott British lamb. \textcolor{Red}{Peter Blackburn} BRUSSELS 1996-08-22 \\
         \midrule
         Peter Blackburn (Bishop) & Peter Blackburn (d.1616) was a Scottish scholar and prelate. He was the second Protestant Bishop of Aberdeen.  \\
         Peter Blackburn (MP) & Peter Blackburn (1811 – 20 May 1870) was a British Conservative Party politician. \\
         Peter Blackburn (Badminton) & Peter Grant Blackburn (born 25 March 1968) is an Australian badminton player who affiliated with the Ballarat Badminton Association.  \\
         \midrule
         \multicolumn{2}{c}{\textbf{Non-Entity Phrase}} \\
         \midrule
         Mention Context & Most Hindus accept that there is a duty to have a family during \textcolor{Red}{the householder} stage of life, as debt to family lineage called Pitra Rin (Father's Debt) and so are unlikely to avoid having children altogether \ldots \\
         \midrule
         The Householder (Film) & The Householder (Hindi title: Gharbar) is a 1963 film by Merchant Ivory Productions, with a screenplay by Ruth Prawer Jhabvala \ldots \\
         The Householder (Novel) & The Householder is a 1960 English-language novel by Ruth Prawer Jhabvala \ldots \\
         \bottomrule
    \end{tabular}
    }
    \caption{Example of mentions that should be linked to \nil{} and their potential candidate entities. Mentions are labeled as \textcolor{Red}{red}.}
    \label{tab:nil_example}
\end{table*}

Previous entity linking datasets have paid insufficient attention to the \nil{} prediction problem.
For example, some of the previous datasets like  AIDA~\cite{hoffart2011robust} view it as an auxiliary task, while others like MSNBC~\cite{cucerzan2007large} and WNED-WIKI~\cite{eshel2017named} does not require \nil{} prediction at all.
There does not yet exist a strong benchmark for the ability on \nil{} prediction.

In this paper, we propose an entity linking dataset \nelname{} that focuses on the \nil{} prediction problem.
About 30\% of the mentions in \nelname{} do not have their corresponding entity in the candidate entity set, which requires models to identify these mentions rather than linking them to the wrong candidates.
In \nelname{} construction, we take ambiguous entities as seeds, and build the dataset by mining mention contexts related to seed entities on the Wikipedia corpus.
Then, human annotators are asked to identify whether the mentions correspond to a candidate entity or not, and we further perform entity masking to ensure a fair proportion of \nil{} data of about 30\%.

In \nil{} prediction, we propose to use the widely used bi-encoder and cross-encoder structures as the model backbone, and further integrate type information by adding an entity typing subtask.
We combine semantic and type similarity as the final similarity score, and identify \nil{} mentions by setting a similarity threshold.

We conduct a series of experiments on both \nelname{} and previous entity linking datasets.
Experimental results show that the models suffer from an accuracy drop when taking \nil{} prediction into consideration, indicating that the accuracy may be inflated without the \nil{} prediction task, and \nelname{} could better diagnose the performance of different models.
We also conducted ablation studies on how type information and \nil{} examples affect the models.
We discover that the entity typing subtask yields better embedding even when type similarity is not used, and both types of \nil{} examples in training data would boost the ability of \nil{} prediction.

Our contributions can be concluded as:
\begin{itemize}
    \item We categorize the \nil{} prediction problem into two patterns: Missing Entity and Non-Entity Phrase, where the latter one has not received sufficient attention in previous works.
    \item We propose an entity linking dataset \nelname{} focusing on \nil{} prediction, which covers two patterns of \nil{} data and could act as a benchmark for diagnosing the ability of \nil{} prediction.
    \item We conducted a series of experiments, whose results demonstrate that the accuracy of models may be inflated when not taking \nil{} prediction into consideration. Meanwhile, both patterns of \nil{} data in training are essential for triggering the ability of \nil{} prediction.
\end{itemize}

\begin{figure*}[htbp]
  \centering
  \includegraphics[width=0.9\linewidth]{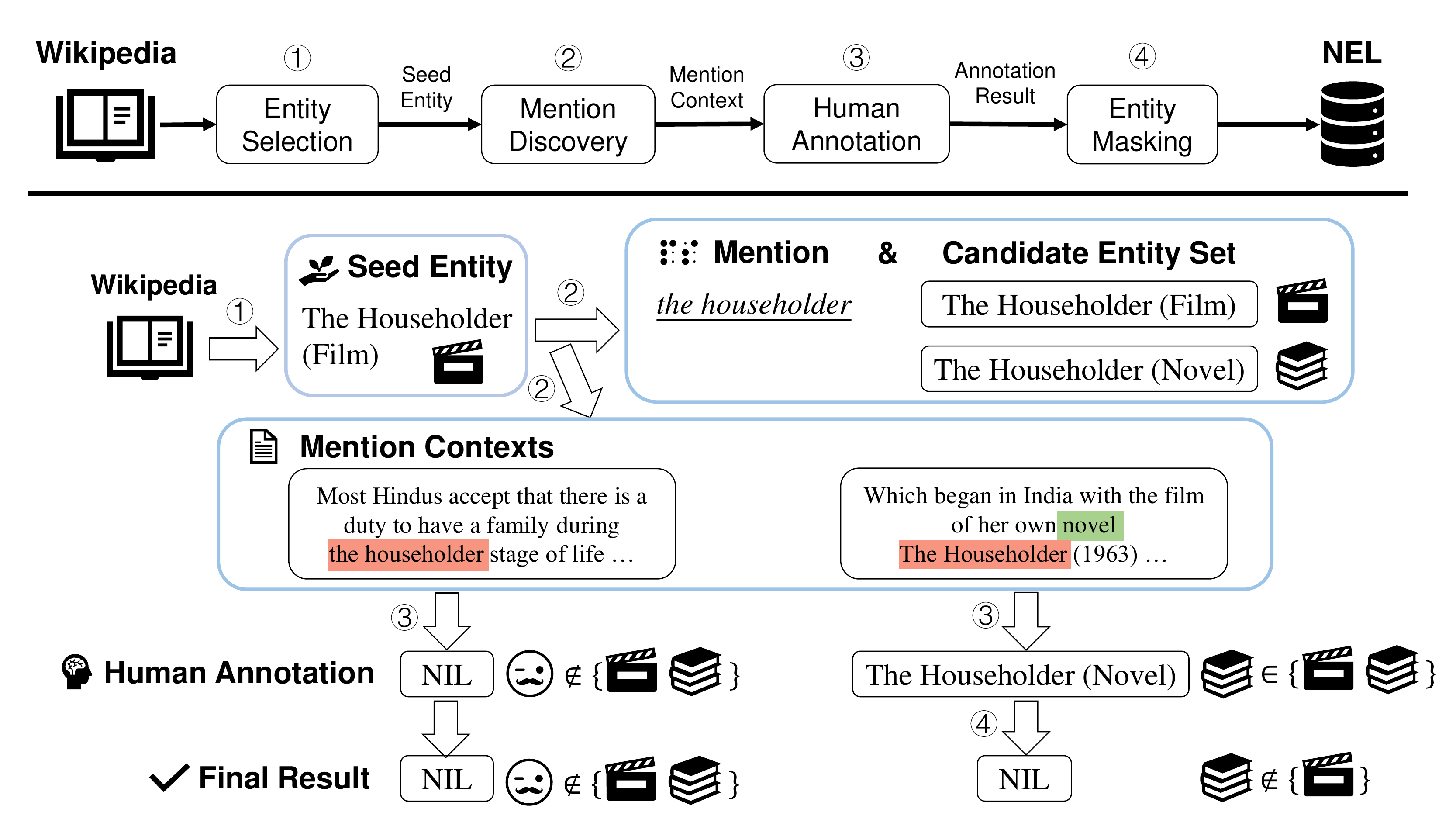}
    \caption{An illustration of how \nelname{} is constructed. We select ambiguous entities as seeds, taking their alias as potential mentions to discover mention contexts from Wikipedia. The entries are annotated by human annotators, and entity masking is performed on some entries to control the portion of \nil{} data.}
  \label{fig:dataset_annotation}
\end{figure*}

\section{Preliminary}
\label{sec:preliminary}
Entity mentions $M = \{m_i\}$ refer to text spans potentially corresponding to entities in a document $D = (w_1, w_2, ..., w_n)$, where $w_i$ is either a plain token or a mention.
Each mention $m_i$ may correspond to an entity $e_i \in E$ in the entity set $E$ of knowledge base $\mathcal{KB}$. 

\begin{definition}
    \label{def:el} 
    \textbf{Entity linking} aims to find an optimal mapping $\Gamma : M \Rightarrow E$, which maps entity mentions $M = \{m_i\}$ to their corresponding entities $E = \{e_i\}$, where $e_i \in \mathcal{KB}$.
\end{definition}

The \nil{} prediction problem is to determine whether an entity mention $m$ is absent from the knowledge base $\mathcal{KB}$.
When there does not exist a proper entity $e \in \mathcal{KB}$ for the given mention $m$, the model should link $m$ to a special \nil{} entity, indicating that the mention is unlinkable.

\begin{definition}
    \label{def:nil_el}
    \textbf{Entity linking with \nil{} prediction} aims to find an optimal mapping $\Gamma : M \Rightarrow E \cup \{NIL\}$, where \nil{} is a special placeholder and $m$ correspond to \nil{} only when there is no correct answer in the candidate entity set $E$.
\end{definition}

As demonstrated in Table \ref{tab:nil_example}, there exist two situations in real-world entity linking where \nil{} prediction should be taken into consideration:
\begin{itemize}
    \item \textbf{Missing Entity:~} The mention $m$ refers to certain entity $e$ that has not been yet included in $\mathcal{KB}$, i.e. $m \Rightarrow e \notin \mathcal{KB}$. For example, in the upper half of Table \ref{tab:nil_example}, the mention \textit{Peter Blackburn} refers to a certain journalist, while entries in English Wikipedia include only people with other occupations, leading to the mention linking to \nil{}.
    \item \textbf{Non-Entity Phrase:~} The mention $m$ refers to a common phrase that is not usually viewed as an entity, i.e. $m \nRightarrow e$. For example, the mention \textit{the householder} in the lower half of Table \ref{tab:nil_example} refers to a concept rather than a film or novel.
\end{itemize}

\begin{table*}[htbp]
    \centering
    \scalebox{0.9}{
    \begin{tabular}{c c c c c c}
        \toprule
        Dataset & \# Data & Annotated & \nil{} percentage & \%Missing Entity & \%Non-Entity Phrase \\
        \midrule
        AIDA & 34956 & \ding{51} & 20.41\% & 73\%* & 10\%*  \\
        MSNBC & 654 & \ding{51} & 0\% & - & - \\
        WNED-Wiki & 240000 & \ding{56} & 0\% & - & -  \\
        \midrule
        \nelname{} (ours) & 9924 & \ding{51} & 33.57\% & 17\% & 83\% \\
        \bottomrule
    \end{tabular}
    }
    \caption{Statistics of the \nelname{} dataset compared with previous entity linking datasets. *The percentage of two \nil{} patterns in AIDA is calculated from 300 randomly sampled \nil{} data, and data with errors do not count as any pattern.}
    \label{tab:nel_statistics}
\end{table*}

\section{Dataset Construction}
There does not yet exist a strong benchmark on \nil{} prediction.
We manually annotated 300 examples with their mentions linking to \nil{} from the widely used entity linking dataset AIDA\footnote{}, and discover that about 10\% of these mentions should actually link to an entity. 
For example, the mention "EU" in "EU rejects German" should be linked to the European Union rather than \nil{} (See Appendix \ref{sec:appendix_AIDA_error} for details).
Meanwhile, \nil{} mentions in AIDA fall mostly in the \textbf{Missing Entity} category.
The incorrect and imbalanced data for \nil{} prediction indicates that the importance of \nil{} prediction is currently underestimated.

In this section, we propose an entity linking dataset named \nelname{}, which focuses on the \nil{} prediction problem.
The construction process is demonstrated in Figure \ref{fig:dataset_annotation}.

Unlike normal entity linking data, there does not exist annotated references for mentions linking to \nil{}, and the portion of \nil{} data in the text corpus is unknown.
Hyperlinks in Wikipedia can be viewed as mentions linking to non-\nil{} entities, from which we can find the aliases of entities.
We assume that when an alias does not appear as a hyperlink in a certain context, it may be identified as a mention linking to \nil{}.
In this way, we collect such contexts as the raw data.
The raw data is then annotated by humans to ensure correctness, and we further mask out some answer entities in the candidate set to control the percentage of \nil{} in answers.

\subsection{Data Collection}
\citet{levin1977title} states that the title of creative works could be a place, a personal name, or a certain abstract concept like the choric embodiment of some collectivity (\textit{The Clouds}, \textit{The Birds}) or stock types (\textit{The Alchemist}, \textit{Le Misanthrope}), which would naturally lead to the two situations where a mention links to \nil{}.
The absence of the referenced entity from the KB would lead to \textbf{Missing Entity}, while an abstract concept not viewed as an entity would lead to \textbf{Non-Entity Phrase}. 

To discover \nil{} mentions of both types, we start by taking entities that share an alias with other entities as seeds.
We assume that a mention referring to multiple entities has a higher probability of linking to a \textbf{Missing Entity} outside the KB, and the complex meaning of the mention will lead to \textbf{Non-Entity Phrase}.
Thus, the aliases of ambiguous seed entities would be good starting points for mining \nil{} mentions.

\paragraph{Entity Selection}
We further filter ambiguous entities to remove low-quality seeds.
First, we remove noise instances like template pages, and entities with less than 5 hyperlinks are also removed. 
Meanwhile, we discarded entities with a probability greater than 50\% of being the linking result, as these entities can generally be considered to be unambiguous and lack difficulty.
Finally, 1000 entities are sampled as the seed entity set $E_s$.

We use a typing system based on Wikidata to identify the type of selected entities.
We view the \textit{instance of} relation as the type indicator, and utilize the \textit{subclass of} relation to build a tree-form type system.
The detailed type system can be found in Appendix \ref{sec:appendix_type}.

\paragraph{Mention Discovery}
We build an alias table from the 2021 Wikipedia dump by extracting alias-entity pairs $(m, e)$ from internal hyperlinks.
All alias $m$ related to entities in the seed entity set $E_s$ are gathered as the mention set $M$.
For each mention $m \in M$, we look for its occurrence throughout the Wikipedia corpus (whether it appears as a hyperlink or not) to obtain the entry tuple $(C_l, m, C_r, E_m)$, where $C_l$ and $C_r$ represent contexts left and right to the entity mention $m$, and $E_m$ represents the candidate entities set of $m$.
For each mention $m$, we sampled 5 entries where $m$ appears as a hyperlink and 5 entries where $m$ appears in plain text to balance the number of positive and negative examples, and a total of 10,000 entries are collected.

\subsection{Human Annotation and Post-processing}
We perform annotation on the above entries with 3 annotators.
The annotators are provided with the mention context $(C_l, m, C_r)$ and candidate entities $E_m$.
Each candidate entity $e$ consists of its title, textual description, and Wikipedia URL.
The annotators are asked to choose the answer entity $a \in E_m$ corresponding to the mention $m$, or $a = NIL$ if none of the candidates are correct.

An expert will further investigate entries in which annotators fail to reach a consensus.
The expert is a senior annotator with essential knowledge of entity linking, and will confirm the final annotation result after carefully reading through the context and candidate entities.
We use the annotation result as the final answer $a$ if there is an agreement between 3 annotators, and the entity chosen by the expert otherwise.
The annotated tuple $(C_l, m, C_r, E_m, a)$ acts as the final entry of our dataset.

To further simulate the situation where new emerging entities do not appear in knowledge bases, we perform entity masking on positive entries.
We randomly sample 10\% entries where $a \neq NIL$, and mask the correct entity in the candidate set $E_m$.
In this case, as the correct answer is removed from the candidate list, we have $a = NIL$, i.e. the mention $m$ corresponds to the empty entity \nil.

\begin{figure*}[htbp]
  \centering
  \includegraphics[width=0.9\linewidth]{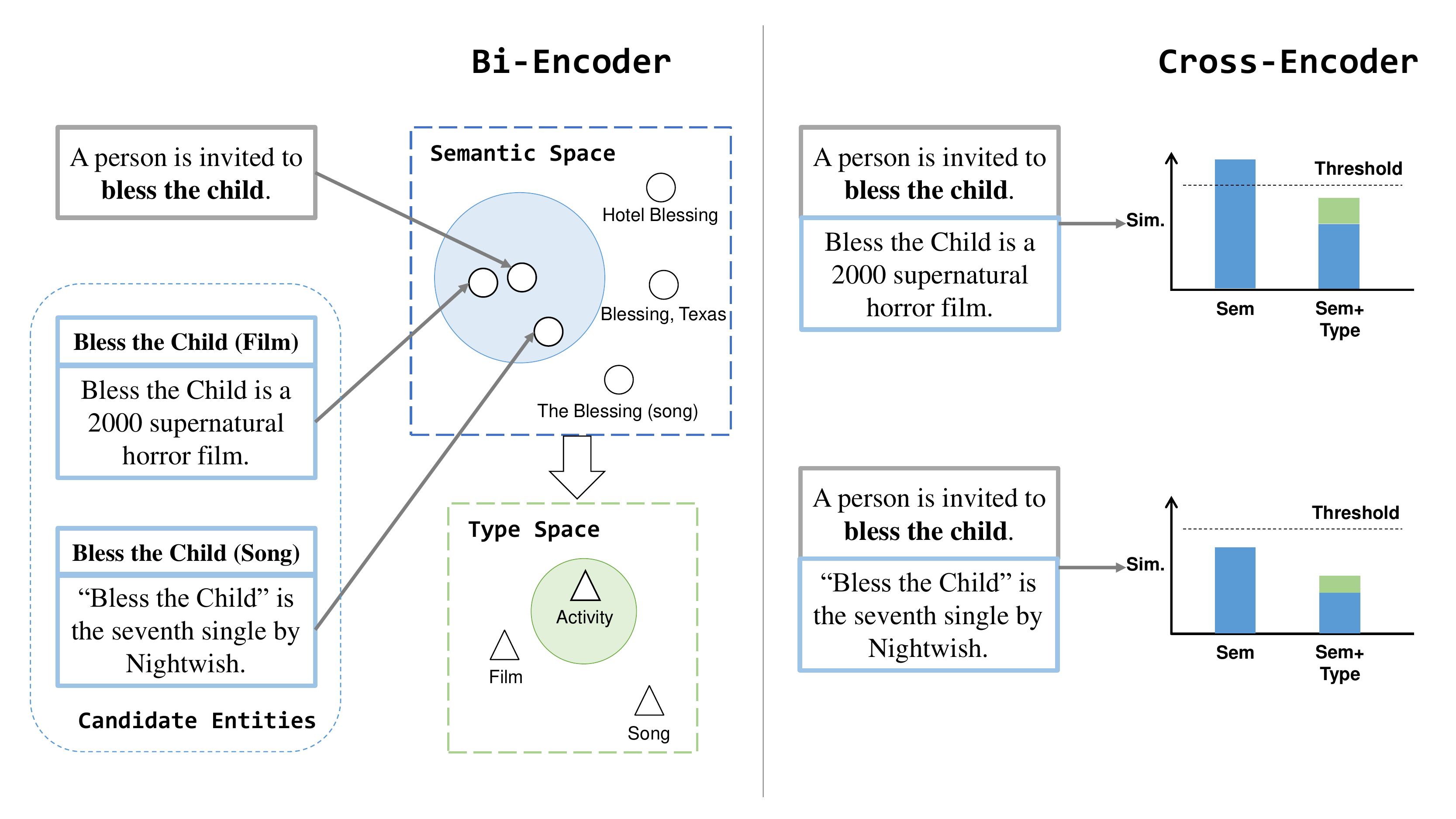}
  \caption{The overall structure of PLM-based retrieval models. Candidates which are confusing in the semantic space may be more distinguishable in the type space. Mentions linking to \nil{} frequently differ from their candidates in their types, so we combine semantic similarity with type similarity for \nil{} prediction.}
  \label{fig:clink_model}
\end{figure*}

\subsection{Dataset Statistics}
Table \ref{tab:nel_statistics} demonstrates the properties of the \nelname{} dataset.
\nelname{} includes 6,593 positive examples and 3,331 negative examples, covering 1,000 mentions and 3,840 related entities.
Each mention has an average of 3.80 candidate entities.
The inter-annotator agreement of \nelname{} is 94.61\%, indicating that the expert calibrated about 5\% of the data. 
The full dataset is split into train/validation/test sets by the ratio of 80\%/10\%/10\%.

\nelname{} contains a fair number of entries, and is human-annotated to ensure correctness.
Compared with previous entity linking datasets, \nelname{} has a higher percentage of \nil{} data, which could better diagnose the ability of different models in \nil{} prediction.
Meanwhile, mentions linking to \nil{} in AIDA mostly fall in the \textbf{Missing Entity} situation, while \nelname{} focuses more on the \textbf{Non-Entity Phrase} situation, thus complementing the insufficient attention on \textbf{Non-Entity Phrase} in \nil{} prediction.

\section{Entity Linking with \nil{} prediction}
A mention links to \nil{} when all of its candidate entities fail to match its context.
A common paradigm of \nil{} prediction is to compute the similarity scores between the mention context and candidates, and judge its linking result on the base of similarities.

\subsection{Scoring Similarities}
Bi-encoder and cross-encoder are widely adopted scorer structures, as they are well compatible with pretrained language models.
Bi-encoder encodes mention contexts and entities into the same dense vector space, while cross-encoder views similarity scoring as a sequence classification task: 
\begin{align}
    s_{bi}(c, e) &= \sigma(f(c) \cdot g(e)) \\
    s_{cross}(c, e) &= \sigma(\textbf{W} h([c,e]) + \textbf{b})
\end{align}
where $f, g, h$ are PLM-based encoders, $\textbf{W}$ and $\textbf{b}$ are trainable variables, and $\sigma$ refers to the sigmoid function.

The bi-encoder structure allows precomputing entity embeddings in knowledge bases, which enables efficient retrieval in real-world applications.
Compared with bi-encoder, cross-encoder better captures the relation between context and entities with the cross attention mechanism, thus demonstrating higher accuracy.

\begin{table*}[htbp]
    \centering
    \caption{Experimental results on \nelname{} and previous datasets. Non-NAC, NAC, and OAC represent non-\nil{} accuracy, \nil{} accuracy and overall accuracy. *Results of GENRE on AIDA w/o \nil{}, MSNBC, and WNED-WIKI are taken from the original paper~\cite{decao2020autoregressive}.}
    \scalebox{0.8}{
    \begin{tabular}{c c c c c c c c c c}
         \toprule
          & \multicolumn{3}{c}{\nelname{} (our dataset)} & \multicolumn{3}{c}{AIDA w/ \nil{}} & AIDA w/o \nil{} & MSNBC & WNED-WIKI \\
          \midrule
          & Non-NAC & NAC & OAC & Non-NAC & NAC & OAC & OAC & OAC & OAC \\
         \midrule
         BLINK-bi & 72.27 & 88.59 & 77.74 & 64.54 & \textbf{69.59} & 65.01 & 82.61 & 70.86 & 58.56 \\  
         \clinkname{}-bi & 79.24 & 79.28 & 79.25 & 75.98 & 66.36 & 75.09 & 83.26 & 73.29 & 58.99 \\
         GENRE* & 54.00 & 62.84 & 56.96 & - & - & - & \textbf{88.60} & 88.10 & 71.70 \\
         BLINK-cross & 84.09 & 88.89 & 85.70 & 83.08 & 45.16 & 79.58 & 87.49 & 82.02 & 69.48 \\
         \clinkname{}-cross & \textbf{86.97} & \textbf{89.19} & \textbf{87.71} & \textbf{84.42} & 58.53 & \textbf{82.03} & 88.16 & \textbf{89.70} & \textbf{72.43} \\
         \bottomrule
    \end{tabular}
    }
    \label{tab:nel_exp_result}
\end{table*}

\subsection{Integrating Entity Types}
Previous entity linking models~\cite{gupta2017entity, onoe2020fine, raiman2018deeptype} have proved that entity types do help models better disambiguate between candidate entities.

The type information can be integrated into bi-encoders and cross-encoders by adding a typing layer.
In the bi-encoder structure, the mention types $t_c$ and entity types $t_e$ are predicted separately:
\begin{align}
    t_c &= \sigma(W_c f(c) + b_c) \\
    t_e &= \sigma(W_e g(e) + b_e)
\end{align}
while they are simultaneously predicted in the cross-encoder structure:
\begin{equation}
    [t_c, t_e] = \sigma(W f([c,e]) + b)
\end{equation}
where $\sigma$ represents the sigmoid function and $W_{c}, b_{c}, W_{e}, b_{e}, W, b$ are trainable parameters.

To tackle the label imbalance between types, we use the focal loss~\cite{lin2017focal} on the typing task:
\begin{equation}
    \mathcal{L}_{t} = -\sum_{i=1}^{n_t} (y_i (1-t_i)^\gamma \log t_i + (1-y_i) t_i^\gamma \log(1-t_i))
\end{equation}
where $n_t$ is the total number of types in the type system, $\gamma$ is a hyperparameter, $y_i$ is the golden label of the $i$-th type and $t_i$ is the predicted label of the $i$-th type.
In bi-encoder, $\mathcal{L}_{t}$ is the average of loss on $t_c$ and $t_e$, while in cross-encoder $\mathcal{L}_{t}$ is directly calculated from $[t_c, t_e]$.

We train the semantic encoder with binary classification loss $\mathcal{L}_s$, and combine $\mathcal{L}_s$ with $\mathcal{L}_{t}$ as the final loss $\mathcal{L}$:
\begin{equation}
    \mathcal{L} = \mathcal{L}_{s} + \mathcal{L}_{t}
\end{equation}

\subsection{Identifying Mentions Linking to \nil{}}
The type similarity is computed with cosine similarity, and the final score is a weighted sum between type similarity and semantic similarity:
\begin{align}
    s_t(c, e) &= cos(t_c, t_e) \\
    s(c, e) &= \lambda s_s(c, e) + (1-\lambda) s_t(c, e)
\end{align}
where $\lambda$ is a hyperparameter.

For each entry $(C_l, m, C_r, E_m, a)$, we concatenate $(C_l, m, C_r)$ to form the context $c$.
In the training step, for each candidate entity $e \in E_m$, we view $(c, e)$ as a positive example if $e = a$, and as a negative example if $a = NIL$ or $e \neq a$.

During evaluation, the similarity score $s(c, e)$ is computed between context $c$ and each candidate entity $e$.
If there exist entities with a score equal to or higher than the nil threshold $\epsilon = 0.5$, we choose the entity with the highest similarity score as the answer; 
If all entities fail to reach the threshold, then the mention $m$ links to \nil.
\begin{equation}
    a = 
    \begin{cases}
        \arg\max_{e} s(c,e),& \exists e, s(c,e) \geq \epsilon \\
        NIL,& \forall e, s(c,e) < \epsilon
    \end{cases}
\end{equation}

\begin{table*}[htbp]
    \centering
    \caption{Experimental results on the influence of the entity typing task on \nelname{}. OAC indicates the overall accuracy of entity linking. The overall accuracy with typing is achieved without using the type similarity score.}
    \begin{tabular}{c c c c c}
        \toprule
        Model & Ctxt Type Acc. & Cand Type Acc. & OAC w/ Typing & OAC w/o Typing \\
        \midrule
        Bi-encoder & 83.75 & 93.46 & 78.35 & 77.74  \\
        Cross-encoder & 83.95 & 98.01 & 87.41 & 83.67  \\
        \bottomrule
    \end{tabular}
    \label{tab:ablation_type_influence}
\end{table*}

\section{Experiments}

We conduct a series of experiments on two types of datasets, which test the different ability of entity linking models: (1) \nelname{} that tests the ability of \nil{} prediction; (2) previous EL datasets that tests the ability of entity disambiguation.
We choose the following models for comparison:
BLINK~\cite{wu2020scalable} that uses the bi-encoder and cross-encoder alone to score candidate entities, \clinkname{} that integrates type information with BLINK, and GENRE~\cite{decao2020autoregressive} that generates the linking result with a sequence-to-sequence model.

\begin{figure*}[htbp]
  \centering
  \includegraphics[width=0.9\linewidth]{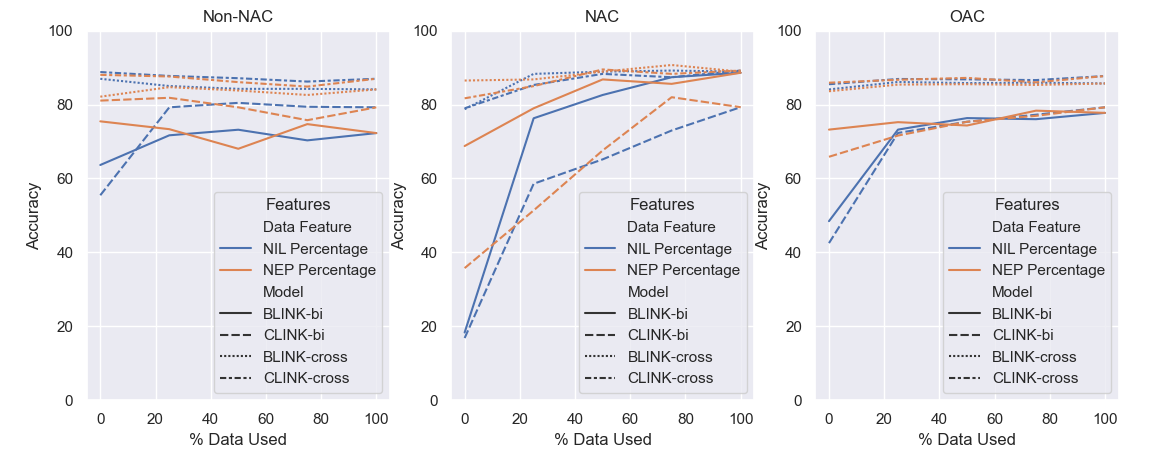}
  \caption{Ablation study on the influence of \nil{} training data. The x-axis indicates the percentage of used \nil{} data or Non-Entity Phrase data in the training set.}
  \label{fig:nil_influence}
\end{figure*}

\subsection{Main Experiment Results}
We trained and tested the models on \nelname{}, with BERT-large as the encoder base of BLINK and \clinkname{}, and BART-large as the backbone of GENRE, to observe their ability in \nil{} prediction.
We also experimented on previous entity linking datasets to observe the disambiguation ability of different models.
The models are trained on the AIDA-YAGO2-train~\cite{hoffart2011robust} dataset, and tested on AIDA-YAGO2-testb, MSNBC~\cite{cucerzan2007large} and WNED-WIKI~\cite{eshel2017named}.

187 distinct types are used in experiments on \nelname{}, and considering that the entity type distribution may be different across datasets, we use a type system with only 14 top-level entity types on previous datasets to make \clinkname{} more transferable (See Appendix \ref{sec:appendix_type} for details).
We retain the same textual representation format with BLINK (see Appendix \ref{sec:appendix_dataset}), while using 128 as the length of context sequences and entity descriptions.
All models are implemented with PyTorch and optimized with the Adam~\cite{kingma2014adam} optimizer with a learning rate of 1e-5.

Table \ref{tab:nel_exp_result} shows the evaluation results, from which some insights could be discovered:
\paragraph{Type Information Matters.} 
\clinkname{} with cross-encoder structure achieves the highest accuracy on almost all datasets, and is still comparable with GENRE on AIDA, from which we may assert that taking type information into consideration is helpful even without \nil{} prediction.
Meanwhile, on both structures, the overall accuracy of \clinkname{} outperforms BLINK on all datasets, proving that the entity typing task assists both bi-encoder and cross-encoder distinguish correct entities.
\paragraph{Encoder Structure.} 
The cross-encoder structure generally performs better than bi-encoder, but we observe that sometimes bi-encoders show decent ability in detecting \nil{} mentions, especially when type information is not utilized.
BLINK-bi achieves the highest \nil{} accuracy score of 69.59 on AIDA with \nil{}, and has a score of 88.59 on \nelname{}, which is comparable with the best-performing CLINK-cross.
This phenomenon indicates that cross-encoders may be more prone to overfitting, while entity types would alleviate this tendency.
\paragraph{\nil{} Entries.}
On the AIDA dataset, we observe that the models generally suffer from a drop in accuracy when taking \nil{} entries into consideration, and the drop is more obvious in bi-encoders.
This may indicate that the performance of models is inflated without \nil{} prediction, and \nil{} entries may confuse the models in practical application.

\subsection{Ablation Study}

\subsubsection{Influence of the Entity Typing Task}
We conducted experiments to observe how the entity typing task influences the model.
We trained the model with both $\mathcal{L}_{t}$ and $\mathcal{L}_{s}$ as loss, while setting $\lambda = 1$ during evaluation to ignore the influence of type similarity scores.
Results shown in Table \ref{tab:ablation_type_influence} reflects two observations:

% First, type predictions on candidate entities benefit more from mention context, possibly because the context helps narrow down the type range, while the context type prediction generally remains unaffected by entity descriptions.
First, candidate type predictions benefit from mention context.
We observe that when changing the structure from bi-encoder to cross-encoder, the type prediction accuracy on candidates raises by 5\%, where the accuracy on contexts remains unchanged.
This is likely because the context helps narrow down the type range, while the context type prediction generally remains unaffected by entity descriptions.

Second, unifying the entity typing task in the training process leads to higher overall accuracy, even when the type similarity score is not taken into consideration in the evaluation, which can be demonstrated by the improved score of OAC w/ Typing compared to OAC w/o Typing on both models.
This may indicate that predicting entity types would help the model learn semantic embeddings with higher quality.

\subsubsection{Influence of \nil{} Training Data}
Compared with previous datasets like AIDA, \nelname{} contains more \nil{} mentions of the \textbf{Non-Entity Phrase} type.
We trained the models with different numbers of \textbf{Non-Entity Phrase} examples to observe the influence of \nil{} Training Data.

As demonstrated by Figure \ref{fig:nil_influence}, all models suffer from a great decline in \nil{} accuracy when no \nil{} examples are used during the training stage, and bi-encoder is more prone to the accuracy drop.
However, by using only 25\% of \textbf{Non-Entity Phrase} examples in training, the \nil{} accuracy would recover to a decent level.
Further adding \nil{} examples has little impact on cross-encoder models, but bi-encoder models still constantly benefit from additional data.

Besides, ignoring \nil{} data with the \textbf{Non-Entity Phrase} type will also harm the \nil{} accuracy and overall accuracy.
Both types of \nil{} training data are necessary to reach to best performance.

We discover that entity linking models may be unaware of the \nil{} mentions when there is insufficient training data.
A small amount of training data is enough for cross-encoder models to reach a reasonable accuracy, while bi-encoder models constantly benefit from additional training data.

\section{Related Work}
\paragraph{PLM-based Models in Entity Linking.~}
Using pretrained language models (PLM) to capture semantic information is widely adopted in recent entity linking models.
BLINK~\cite{wu2020scalable} marks the mention in context with pre-defined special tokens, and takes BERT as the encoder base.
Two structures are adopted by BLINK to handle different situations: bi-encoder for fast dense retrieval, and cross-encoder for further disambiguation.
MOLEMAN~\cite{fitzgerald2021moleman} searches for similar mention contexts instead of entities, which better captures the diverse aspects an entity reflects in various contexts.
GENRE~\cite{decao2020autoregressive} finetunes the sequence-to-sequence model BART, directly generating the unique entity name according to the mention context.

\paragraph{Research on \nil{} Prediction.~}
The \nil{} prediction problem has been long viewed as an auxiliary task of entity linking.
Some entity linking datasets (AIDA~\cite{hoffart2011robust}, TAC-KBP series~\cite{mcnamee2009overview}) take the \nil{} prediction problem into consideration, while some (ACE and MSNBC)~\cite{ratinov2011local} omit mentions linking to \nil.
Some research has already been conducted on the \nil{} prediction problem.
\citet{lazic2015plato} and \citet{peters2019knowledge} set a score threshold to filter reasonable candidates, and mentions with no candidate score above the threshold are linked to \nil{}.
\citet{sil2016one, kolitsas2018end} views the \nil{} placeholder as a special entity, and selecting it as the best match indicates that the mention refers to no entities in the given KB.
However, recent entity linking models, which use pretrained language models (PLM) as encoder bases, generally take the in-KB setting, which assumes that each mention has a valid golden entity in the KB~\cite{wu2020scalable}.

\paragraph{Entity Type Assisted Entity Linking.~}
Entity types can effectively assist entity linking and have been studied in various works.
\citet{gupta2017entity} jointly encodes mention context, entity description, and Freebase types with bidirectional LSTM to maximize the cosine similarity.
DeepType~\cite{raiman2018deeptype} predicts the type probability of each token and gathers relevant tokens to predict the entity types, which would help eliminate candidates with incompatible types.
\citet{onoe2020fine} views entity types as a training objective rather than a feature, predicting fine-grained Wikipedia category tags to select the most relevant entity.

\section{Conclusion}
In this paper, we propose an entity linking dataset \nelname{} that focuses on the \nil{} prediction problem.
We observe that mentions linking to \nil{} can be categorized into two patterns: \textbf{Missing Entity} and \textbf{Non-Entity Phrase}, but the latter one has not been paid sufficient attention.
We propose an entity linking dataset \nelname{} that focuses on \nil{} prediction.
The dataset is built upon the Wikipedia corpus by choosing ambiguous entities as seeds and collecting relevant mention contexts.
\nelname{} is human-annotated to ensure correctness, and
entity masking is further performed to control the percentage of \nil{}.

We conducted a series of experiments to examine the performance of PLM-based models on different datasets.
Experimental results indicate that the accuracy without considering \nil{} prediction would be inflated.
Meanwhile, sufficient data of both \nil{} types during training is essential to trigger the ability of \nil{} prediction.
In the future, we may further try to integrate entity types into the pretraining process and explore type transfer between datasets.

% \clearpage

\section*{Acknowledgements}
We thank the anonymous reviewers for their insightful comments. 
This paper is supported by a grant from the Institute for Guo Qiang, Tsinghua University (2019GQB0003).

\section*{Limitations}
Our work still exist some limitations.
First, we choose an entity typing system on the base of Wikidata tags, however, the granularity of the typing system remains to be discussed.
A system with too many types would introduce noise to long-tail types, while insufficient types would weaken the disambiguation ability of type similarity.
Thus, building a type system with adequate granularity remains a challenge.

Second, we combine the entity typing task with PLM-based semantic encoders, which require a fixed type system and further finetuning.
Integrating the entity typing task into the pretraining process may enhance the transferability of the model and remove the dependency on a fixed type system.

\paragraph{Potential Risks.}
Our proposed dataset \nelname{} centers on ambiguous entities, whose type distribution may not remain the same with other datasets.
A potential risk is that the model trained on \nelname{} may experience under-exposure of other entity types, which would damage their transferability and lead to undesired outputs on other datasets.

\section*{Ethics Statement}
In this section, we will discuss the ethical considerations of our work.
\paragraph{Licenses and terms.}
The Wikipedia corpus and Wikidata types are obtained via the Wikimedia dump\footnote{https://dumps.wikimedia.org}, under the CC BY-SA 3.0 license\footnote{https://creativecommons.org/licenses/by-sa/3.0/}.
AIDA, MSNBC, and WNED-WIKI are shared under the CC BY-SA 3.0 license.
These datasets have been widely used in entity linking research, and we believe that they have been anonymized and desensitized.
\paragraph{Human Annotation.}
We recruited 3 human annotators without a background of expertise in annotation, and 1 expert annotator with adequate knowledge in entity linking for checking.
These annotators are employed by commercial data annotation companies.
We have paid these recruited annotators with adequate rewards under the agreed working time and price.
The annotators are well informed about how these annotated data will be used and released, which has been recorded in the contract.
\paragraph{Intended use.}
\nelname{} is an entity linking dataset focusing on the \nil{} prediction problem.
Researchers are intended to use \nelname{} for examining the ability of \nil{} prediction of newly created entity linking models.
AIDA, MSNBC, and WNED-WIKI are intended for entity linking research, which is compatible with our work.

\bibliography{anthology,custom}

\begin{thebibliography}{22}
\expandafter\ifx\csname natexlab\endcsname\relax\def\natexlab#1{#1}\fi

\bibitem[{Cao et~al.(2021)Cao, Izacard, Riedel, and
  Petroni}]{decao2020autoregressive}
Nicola~De Cao, Gautier Izacard, Sebastian Riedel, and Fabio Petroni. 2021.
\newblock \href {https://openreview.net/forum?id=5k8F6UU39V} {Autoregressive
  entity retrieval}.
\newblock In \emph{International Conference on Learning Representations}.

\bibitem[{Chen et~al.(2021)Chen, Varoquaux, and Suchanek}]{chen2021lightweight}
Lihu Chen, Ga{\"e}l Varoquaux, and Fabian~M Suchanek. 2021.
\newblock A lightweight neural model for biomedical entity linking.
\newblock In \emph{Proceedings of the AAAI Conference on Artificial
  Intelligence}, volume~35, pages 12657--12665.

\bibitem[{Cucerzan(2007)}]{cucerzan2007large}
Silviu Cucerzan. 2007.
\newblock Large-scale named entity disambiguation based on wikipedia data.
\newblock In \emph{Proceedings of the 2007 joint conference on empirical
  methods in natural language processing and computational natural language
  learning (EMNLP-CoNLL)}, pages 708--716.

\bibitem[{Dredze et~al.(2010)Dredze, McNamee, Rao, Gerber, and
  Finin}]{dredze2010entity}
Mark Dredze, Paul McNamee, Delip Rao, Adam Gerber, and Tim Finin. 2010.
\newblock Entity disambiguation for knowledge base population.
\newblock In \emph{Proceedings of the 23rd International Conference on
  Computational Linguistics (Coling 2010)}, pages 277--285.

\bibitem[{Eshel et~al.(2017)Eshel, Cohen, Radinsky, Markovitch, Yamada, and
  Levy}]{eshel2017named}
Yotam Eshel, Noam Cohen, Kira Radinsky, Shaul Markovitch, Ikuya Yamada, and
  Omer Levy. 2017.
\newblock Named entity disambiguation for noisy text.
\newblock In \emph{Proceedings of the 21st Conference on Computational Natural
  Language Learning (CoNLL 2017)}, pages 58--68.

\bibitem[{Fitzgerald et~al.(2021)Fitzgerald, Bikel, Botha, Gillick,
  Kwiatkowski, and McCallum}]{fitzgerald2021moleman}
Nicholas Fitzgerald, Dan Bikel, Jan Botha, Dan Gillick, Tom Kwiatkowski, and
  Andrew McCallum. 2021.
\newblock Moleman: Mention-only linking of entities with a mention annotation
  network.
\newblock In \emph{Proceedings of the 59th Annual Meeting of the Association
  for Computational Linguistics and the 11th International Joint Conference on
  Natural Language Processing (Volume 2: Short Papers)}, pages 278--285.

\bibitem[{Gupta et~al.(2017)Gupta, Singh, and Roth}]{gupta2017entity}
Nitish Gupta, Sameer Singh, and Dan Roth. 2017.
\newblock Entity linking via joint encoding of types, descriptions, and
  context.
\newblock In \emph{Proceedings of the 2017 Conference on Empirical Methods in
  Natural Language Processing}, pages 2681--2690.

\bibitem[{Hoffart et~al.(2011)Hoffart, Yosef, Bordino, F{\"u}rstenau, Pinkal,
  Spaniol, Taneva, Thater, and Weikum}]{hoffart2011robust}
Johannes Hoffart, Mohamed~Amir Yosef, Ilaria Bordino, Hagen F{\"u}rstenau,
  Manfred Pinkal, Marc Spaniol, Bilyana Taneva, Stefan Thater, and Gerhard
  Weikum. 2011.
\newblock Robust disambiguation of named entities in text.
\newblock In \emph{Proceedings of the 2011 conference on empirical methods in
  natural language processing}, pages 782--792.

\bibitem[{Kingma and Ba(2014)}]{kingma2014adam}
Diederik~P Kingma and Jimmy Ba. 2014.
\newblock Adam: A method for stochastic optimization.
\newblock \emph{arXiv preprint arXiv:1412.6980}.

\bibitem[{Kolitsas et~al.(2018)Kolitsas, Ganea, and Hofmann}]{kolitsas2018end}
Nikolaos Kolitsas, Octavian-Eugen Ganea, and Thomas Hofmann. 2018.
\newblock End-to-end neural entity linking.
\newblock In \emph{Proceedings of the 22nd Conference on Computational Natural
  Language Learning}, pages 519--529.

\bibitem[{Lazic et~al.(2015)Lazic, Subramanya, Ringgaard, and
  Pereira}]{lazic2015plato}
Nevena Lazic, Amarnag Subramanya, Michael Ringgaard, and Fernando Pereira.
  2015.
\newblock Plato: A selective context model for entity resolution.
\newblock \emph{Transactions of the Association for Computational Linguistics},
  3:503--515.

\bibitem[{Levin(1977)}]{levin1977title}
Harry Levin. 1977.
\newblock The title as a literary genre.
\newblock \emph{The Modern language review}, 72(4):xxiii--xxxvi.

\bibitem[{Lin et~al.(2017)Lin, Goyal, Girshick, He, and
  Doll{\'a}r}]{lin2017focal}
Tsung-Yi Lin, Priya Goyal, Ross Girshick, Kaiming He, and Piotr Doll{\'a}r.
  2017.
\newblock Focal loss for dense object detection.
\newblock In \emph{Proceedings of the IEEE international conference on computer
  vision}, pages 2980--2988.

\bibitem[{Ling et~al.(2015)Ling, Singh, and Weld}]{ling2015design}
Xiao Ling, Sameer Singh, and Daniel~S Weld. 2015.
\newblock Design challenges for entity linking.
\newblock \emph{Transactions of the Association for Computational Linguistics},
  3:315--328.

\bibitem[{Luo et~al.(2018)Luo, Lin, Luo, and Zhu}]{luo2018knowledge}
Kangqi Luo, Fengli Lin, Xusheng Luo, and Kenny Zhu. 2018.
\newblock Knowledge base question answering via encoding of complex query
  graphs.
\newblock In \emph{Proceedings of the 2018 Conference on Empirical Methods in
  Natural Language Processing}, pages 2185--2194.

\bibitem[{McNamee and Dang(2009)}]{mcnamee2009overview}
Paul McNamee and Hoa~Trang Dang. 2009.
\newblock Overview of the tac 2009 knowledge base population track.
\newblock In \emph{Text analysis conference (TAC)}, volume~17, pages 111--113.

\bibitem[{Onoe and Durrett(2020)}]{onoe2020fine}
Yasumasa Onoe and Greg Durrett. 2020.
\newblock Fine-grained entity typing for domain independent entity linking.
\newblock In \emph{Proceedings of the AAAI Conference on Artificial
  Intelligence}, volume~34, pages 8576--8583.

\bibitem[{Peters et~al.(2019)Peters, Neumann, Logan, Schwartz, Joshi, Singh,
  and Smith}]{peters2019knowledge}
Matthew~E Peters, Mark Neumann, Robert Logan, Roy Schwartz, Vidur Joshi, Sameer
  Singh, and Noah~A Smith. 2019.
\newblock Knowledge enhanced contextual word representations.
\newblock In \emph{Proceedings of the 2019 Conference on Empirical Methods in
  Natural Language Processing and the 9th International Joint Conference on
  Natural Language Processing (EMNLP-IJCNLP)}, pages 43--54.

\bibitem[{Raiman and Raiman(2018)}]{raiman2018deeptype}
Jonathan Raiman and Olivier Raiman. 2018.
\newblock Deeptype: multilingual entity linking by neural type system
  evolution.
\newblock In \emph{Proceedings of the AAAI Conference on Artificial
  Intelligence}, volume~32.

\bibitem[{Ratinov et~al.(2011)Ratinov, Roth, Downey, and
  Anderson}]{ratinov2011local}
Lev Ratinov, Dan Roth, Doug Downey, and Mike Anderson. 2011.
\newblock Local and global algorithms for disambiguation to wikipedia.
\newblock In \emph{Proceedings of the 49th annual meeting of the association
  for computational linguistics: Human language technologies}, pages
  1375--1384.

\bibitem[{Sil and Florian(2016)}]{sil2016one}
Avirup Sil and Radu Florian. 2016.
\newblock One for all: Towards language independent named entity linking.
\newblock In \emph{Proceedings of the 54th Annual Meeting of the Association
  for Computational Linguistics (Volume 1: Long Papers)}, pages 2255--2264.

\bibitem[{Wu et~al.(2020)Wu, Petroni, Josifoski, Riedel, and
  Zettlemoyer}]{wu2020scalable}
Ledell Wu, Fabio Petroni, Martin Josifoski, Sebastian Riedel, and Luke
  Zettlemoyer. 2020.
\newblock Scalable zero-shot entity linking with dense entity retrieval.
\newblock In \emph{Proceedings of the 2020 Conference on Empirical Methods in
  Natural Language Processing (EMNLP)}, pages 6397--6407.

\end{thebibliography}
\bibliographystyle{acl_natbib}

\appendix

\section{Details about the \nelname{} Dataset Construction}
\label{sec:appendix_dataset}

\begin{table*}[htbp]
    \centering
    \begin{tabular}{|l|c|c|c|c|}
         \hline
          & \multicolumn{4}{c|}{\textbf{NEL} \& AIDA w/ NIL} \\
         \cline{2-5}
          & \multicolumn{2}{c|}{\clinkname{}-bi} & \multicolumn{2}{c|}{\clinkname{}-cross} \\
         \cline{2-5}
         Hyperparameter & Value & Range & Value & Range \\
         \hline
         Learning Rate & 1e-5 & \{1e-3, 1e-5, 3e-5\} & 1e-5 & \{1e-3, 1e-5, 3e-5\} \\
         \hline
         $\lambda$ & 0.5 & & 0.5 & \\
         \hline
         $\epsilon$ & 0.5 & & 0.5 & \\
         \hline
         Epoch & 4 & \{1, 4\} & 4 & \{1, 4\} \\
         \hline
         Batch Size & 4 & \{1, 4, 8, 16, 32\} & 1 & \{1, 4, 8, 16, 32\} \\
         \hline
         \# Parameters & 670M & - & 335M & - \\
         \hline
         Training Time & \~2 hrs & - & \~3.5 hrs & - \\
         \hline
          & \multicolumn{4}{c|}{\textbf{AIDA w/o NIL, MSNBC \& WNED-WIKI}} \\
         \cline{2-5}
          & \multicolumn{2}{c|}{\clinkname{}-bi} & \multicolumn{2}{c|}{\clinkname{}-cross} \\
         \cline{2-5}
         Hyperparameter & Value & Range & Value & Range \\
         \hline
         Learning Rate & 1e-5 & \{1e-3, 1e-5, 3e-5\} & 1e-5 & \{1e-3, 1e-5, 3e-5\} \\
         \hline
         $\lambda$ & 0.9 & [0.0, 1.0] & 0.9 & [0.0, 1.0] \\
         \hline
         $\epsilon$ & 0 & - & 0 & - \\
         \hline
         Epoch & 1 & \{1, 4\} & 1 & \{1, 4\} \\
         \hline
         Batch Size & 4 & \{1, 4, 8, 16, 32\} & 1 & \{1, 4, 8, 16, 32\} \\
         \hline
         \# Parameters & 670M & - & 335M & - \\
         \hline
         Training Time & \~4 hrs & - & \~3 hrs & - \\
         \hline
    \end{tabular}
    \caption{The hyperparameters used in the training process.}
    \label{tab:appendix_hyperparameters}
\end{table*}

\subsection{Corpora} 
The \nelname{} dataset is built from the 2021-07 English Wikipedia dump under the CC BY-SA 3.0 license.
We take the hyperlinks in the raw xml dump as entity mentions, and retain at most 128 tokens around the mentions as their context.
We take 64 tokens left to the mention and 64 tokens right to the mention by default, and more tokens will be included in one side if the other side does not contain enough tokens.
We then discard tokens from both ends to ensure that the context plus the mention do not exceed the 128 token limit.
Media files (image, audio) and Lua commands are discarded during preprocess.

\subsection{Data Selection}
Entries with the following features are viewed as noise and discarded:
\begin{itemize}
    \item The mention context contains the token '*', which is usually a list or formula;
    \item The mention with only 1 candidate entity, which does not pose much challenge;
    \item The mention with more than 20 candidate entities, which is far too challenging;
    \item The mention appears as a sub-span of a word;
    \item The mention has a probability of over 50\% of linking to a certain candidate entity, in which case we view the mention as unambiguous.
\end{itemize}

\subsection{Textual Representation Format}
Textual representation format for bi-encoder:
\begin{align*}
    C &= [\text{CLS}]\ C_l\ [m_{start}]\ m\ [m_{end}]\ C_r\ [\text{SEP}] \\
    E &= [\text{CLS}]\ e_{title}\ [m_{title}]\ e_{desc}\ [\text{SEP}] 
\end{align*}

Textual representation format for cross-encoder:
\begin{align*}
    (C, E) &= [\text{CLS}]\ C_l\ [m_{start}]\ m\ [m_{end}]\ C_r\\ &\ [\text{SEP}]\ e_{title}\ [m_{title}]\ e_{desc}\ [\text{SEP}]
\end{align*}
where $C$ represents mention context and $E$ represents the textual description of candidate mentions.
$[m_{start}], [m_{end}], [m_{title}]$ are special tokens.

\section{Experiment Details}
\label{sec:appendix_exp}
We use the BERT-large-uncased model as the encoder base, with parameters initialized from the python \textit{transformers} library.
The models are trained on a single NVIDIA GeForce RTX 3090 GPU.
We obtain the AIDA, MSNBC and WNED-WIKI dataset from the BLINK~\cite{wu2020scalable} repository~\url{https://github.com/facebookresearch/BLINK}.
We trained our model on the AIDA-train split, and evaluated on all three datasets.

The hyperparameter configurations are as follows.
Detailed hyperparameters are shown in Table \ref{tab:appendix_hyperparameters}.

\begin{table*}[htbp]
    \centering
    \caption{Examples of type lines}
    \begin{tabular}{l p{0.6\linewidth}}
        \toprule
        Entity & Types \\
        \midrule
        14th Street (Manhattan) & Road->RouteOfTransportation->Infrastructure->ArchitecturalStructure->Place \\  
        1958 Copa del Generalísimo & SoccerTournament->Tournament->SportsEvent->SocietalEvent->Event \\
        ATM (song) & Song->MusicalWork->Work \\
        Brats (1991 film) & Film->Work \\
        Babe Ruth & BaseballPlayer->Athlete->Person \\
        \bottomrule
    \end{tabular}
    \label{tab:appendix_type_lines}
\end{table*}

\section{Typing System}
\label{sec:appendix_type}

\subsection{Typing System on \nelname{}}
We use a tree-like typing system with 187 distinct types on \nelname{}.
The typing system is build on the base of Wikidata types.
Table \ref{tab:appendix_type_lines} shows some examples of type lines in the system.
The most 10 frequent types in \nelname{} are: (Work, Organisation, Place, Event, Person, Activity, FictionalCharacter, Award, Species, MeanOfTransportation).

\subsection{Typing System on Traditional Datasets}
We retain 14 top-level types to make \clinkname{} more transferable on different datasets.
These types are: (Other, Person, Place, Work, Organization, Event, Fictional Character, Species, Activity, Device, Topical Concept, Ethnic Group, Food, Disease)

\begin{table*}[htbp]
    \centering
    \scalebox{0.86}{
    \begin{tabular}{p{0.7\textwidth}|l|l}
        \toprule
        Mention Context & Mention & Assumed Entity \\
        \midrule
        Bosnian premier in \textcolor{Red}{Turkey} for one day visit .  ANKARA 1996-08-27 & Turkey & Turkey (Country) \\
        \midrule
        \textcolor{Red}{EU} rejects German call to boycott British lamb .  Peter Blackburn  BRUSSELS 1996-08-22 & EU & European Union \\
        \midrule
        U.S. \textcolor{Red}{F-14} catches fire while landing in Israel .  JERUSALEM 1996-08-25  A U.S. fighter plane blew a tyre and \ldots & F-14 & Grumman F-14 Tomcat \\
        \midrule
        This is the leading story in the Mozambican press on Monday. \textcolor{Red}{Reuters} has not verified this story and does not vouch for its accuracy. & Reuters & Reuters \\
        \bottomrule
    \end{tabular}
    }
    \caption{Examples data in AIDA that are incorrectly linked to \nil{}.}
    \label{tab:appendix_AIDA_error}
\end{table*}

\section{Errors in AIDA}
\label{sec:appendix_AIDA_error}
Table \ref{tab:appendix_AIDA_error} demonstrates some mentions in AIDA that are incorrectly linked to \nil{}.
50 errors are detected among the 300 randomly sampled data in AIDA.

\begin{table*}[htbp]
    \centering
    \scalebox{0.86}{
    \begin{tabular}{l|llll}
        \toprule
        \multirow{3}{*}{\textbf{Mention Context:}}
        & \multicolumn{4}{l}{
        \ldots singer, Michelle Branch, during her visit in Japan to promote her album Hotel Paper, for 
        } \\
        & \multicolumn{4}{l}{
        a magazine interview and photoshoot. After the release of \textcolor{Red}{Gates of Heaven} the group a short 
        } \\
        & \multicolumn{4}{l}{
        break and performed in New York City \ldots
        } \\
        \midrule
        \textbf{Model} & BLINK-bi & CLINK-bi & BLINK-cross & CLINK-cross \\
        \midrule
        \textbf{Prediction} & Gates of Heaven & Gates of Heaven & \nil{} & \textbf{Gates of Heaven (album)} \\
        \bottomrule
        \toprule
        \multirow{3}{*}{\textbf{Mention Context:}}
        & \multicolumn{4}{l}{
        \ldots Michael Williams and David Collins founded the company in 1994 focusing on independent
        } \\
        & \multicolumn{4}{l}{
        features, including Never Met Picasso (1996), \textcolor{Red}{Home Before Dark} (1997), Six Ways To Sunday 
        } \\
        & \multicolumn{4}{l}{
        (1998), \ldots
        } \\
        \midrule
        \textbf{Model} & BLINK-bi & CLINK-bi & BLINK-cross & CLINK-cross \\
        \midrule
        \textbf{Prediction} & Home Before Dark & Home Before Dark & Home Before Dark & \textbf{NIL} \\
        \bottomrule
    \end{tabular}
    }
    \caption{Examples of predicted entity by different models. Entity mentions in context are labelled as \textcolor{Red}{red} and the correct answer is labelled as \textbf{bold}.}
    \label{tab:case_study}
\end{table*}

\section{Case Study}
Table \ref{tab:case_study} shows some examples predicted by \clinkname{} and BLINK without type information, which reflects how entity types influence the linking result.

In the first example, models with the bi-encoder structure incorrectly take the "Gates of Heaven" entry (which is in fact a documentary film) as the linking result, while \clinkname{}-cross notices the context word "album" may indicate the entity type type, and correctly links the mention to the album.
In the second example, the "Home Before Dark" mention actually refer to a 1997 movie\footnote{https://www.imdb.com/title/tt0116547/} like other movies in the context, however the corresponding entry is absent in the English Wikipedia. 
The \clinkname{}-cross model is able to identify that the mention should be labelled as \nil{}, where the other models mistakenly link it to the "Home Before Dark" entry, which is an album rather than a movie.
We observe that the entity types do help measure the similarity between mentions and entities, which enhances the performance of \clinkname{}.

\end{document}